\documentclass[sigconf]{acmart}
\usepackage[noend]{algpseudocode}
\usepackage{algorithmicx,algorithm}
\AtBeginDocument{%
  \providecommand\BibTeX{{%
    \normalfont B\kern-0.5em{\scshape i\kern-0.25em b}\kern-0.8em\TeX}}}

\copyrightyear{2020}
\acmYear{2020}
\setcopyright{acmcopyright}
\acmConference[MM '20]{Proceedings of the 28th ACM International Conference on Multimedia}{October 12--16, 2020}{Seattle, WA, USA}
\acmBooktitle{Proceedings of the 28th ACM International Conference on Multimedia (MM '20), October 12--16, 2020, Seattle, WA, USA}
\acmPrice{15.00}
\acmDOI{10.1145/3394171.3414005}
\acmISBN{978-1-4503-7988-5/20/10}
\acmSubmissionID{787}

\begin{document}

\title{ChoreoNet: Towards Music to Dance Synthesis with Choreographic Action Unit}

%% Remove header for camera-ready version.
\fancyhead{}

\author{Zijie Ye}
\authornote{Key Laboratory of Pervasive Computing, Ministry of Education.}
\authornote{Beijing National Research Center for Information Science and Technology.}
\email{yzjscwy@gmail.com}
\affiliation{
  \institution{Department of Computer Science and Technology, Tsinghua University, Beijing 100084, China}
}

\author{Haozhe Wu}
\authornotemark[1]
\authornotemark[2]
\email{wuhz19@mails.tsinghua.edu.cn}
\affiliation{
  \institution{Department of Computer Science and Technology, Tsinghua University, Beijing 100084, China}
}

\author{Jia Jia}
\authornotemark[1]
\authornotemark[2]
\authornote{Corresponding author.}
\email{jjia@tsinghua.edu.cn}
\affiliation{
  \institution{Department of Computer Science and Technology, Tsinghua University, Beijing 100084, China}
}

\author{Yaohua Bu}
\authornotemark[1]
\authornotemark[2]
\email{18610082539@163.com}
\affiliation{
  \institution{Department of Computer Science and Technology, Tsinghua University, Beijing 100084, China}
}

\author{Wei Chen}
\email{chenweibj8871@sogou-inc.com}
\affiliation{
  \institution{Sogou, Inc.}
}

\author{Fanbo Meng}
\email{mengfanbosi0935@sogou-inc.com}
\affiliation{
  \institution{Sogou, Inc.}
}

\author{Yanfeng Wang}
\email{wangyanfeng@sogou-inc.com}
\affiliation{
  \institution{Sogou, Inc.}
}

\renewcommand{\shortauthors}{Ye, et al}

\begin{abstract}
  Dance and music are two highly correlated artistic forms. Synthesizing dance motions has attracted much attention recently.
  Most previous works conduct music-to-dance synthesis via directly music to human skeleton keypoints mapping. Meanwhile, human
  choreographers design dance motions from music in a two-stage manner: they firstly devise multiple choreographic dance units
  (CAUs), each with a series of dance motions, and then arrange the CAU sequence according to the rhythm, melody and emotion of
  the music. Inspired by these, we systematically study such two-stage choreography approach and construct a dataset to incorporate such choreography
  knowledge. Based on the constructed dataset, we design a two-stage music-to-dance synthesis framework ChoreoNet to imitate human
  choreography procedure. Our framework firstly devises a CAU prediction model to learn the mapping relationship between
  music and CAU sequences. Afterwards, we devise a spatial-temporal inpainting model to convert the CAU sequence into continuous dance
  motions. Experimental results demonstrate that the proposed ChoreoNet outperforms baseline methods ($\mathbf{0.622}$ in terms of CAU BLEU score
  and $\mathbf{1.59}$ in terms of user study score).
\end{abstract}




\maketitle

\section{Introduction}

A famous choreographer named Martha Graham once said, 'Dance is the hidden language of the soul'. As an artistic form with a long history,
dance is an important medium for people to express their feelings. Conventionally, dance is always accompanied by music. Dancers start
to dance when the musical atmosphere is going up at the beginning, perform different dance actions according to the rhythm, melody and
emotion of music clips, and take a bow at the end. The complicated mapping relationship between dance and music has prompted
researchers to investigate dance-to-music synthesis automatically.

\begin{figure}[!t]
  \centering
  \includegraphics[width=0.6\linewidth]{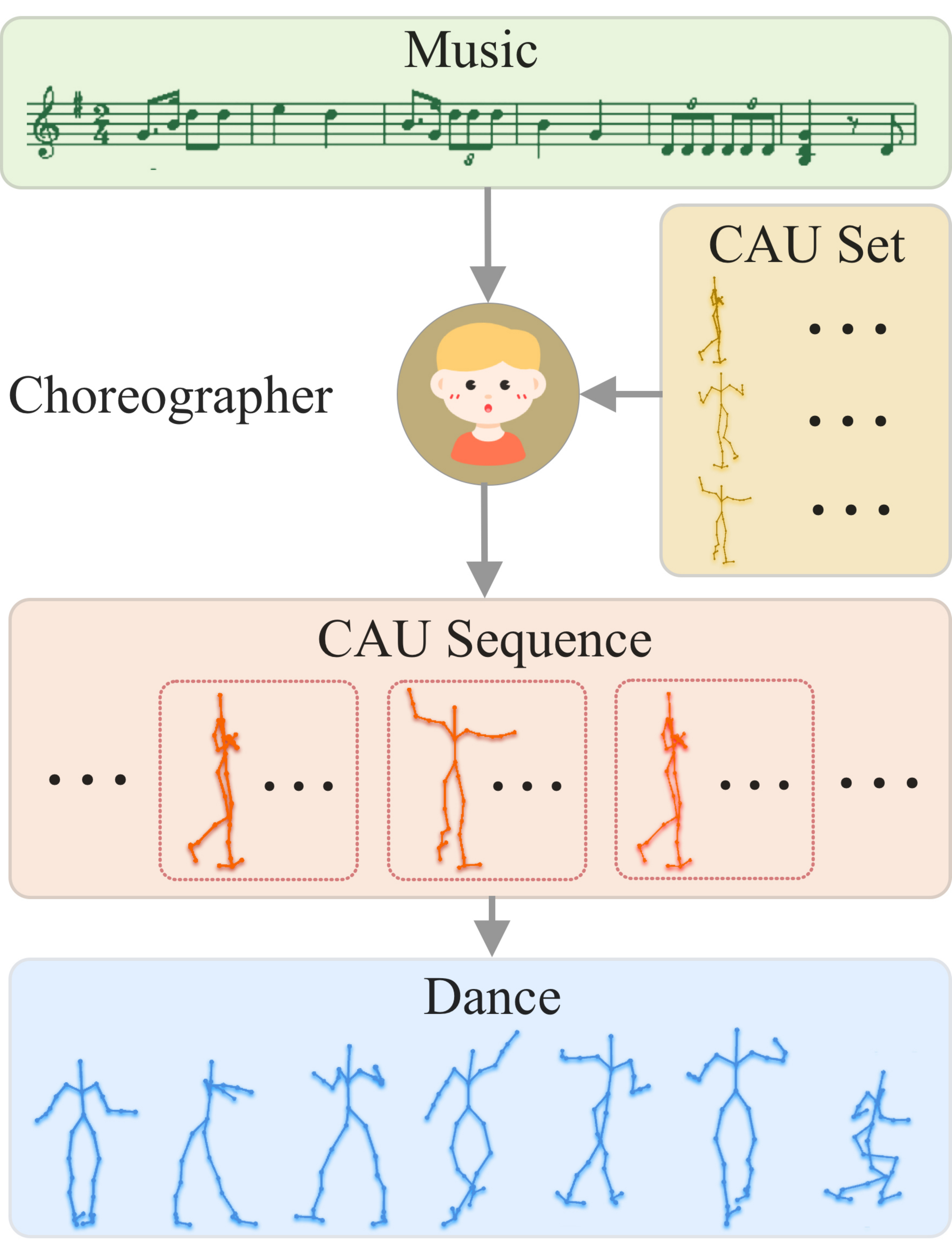}
  \caption{Human choreography procedure: firstly recollect CAUs and then arrange CAUs according to the rhythm, melody and emotion of the music. }
  \label{fig:motivation}
\end{figure}

Several previous research efforts have shown the rationality of music-to-dance synthesis~\cite{lee2013music,cardle2002music,lee2018listen,tang2018dance,lee2019dancing}.
Early works conduct music-to-dance synthesis via solving a similarity-based retrieval problem~\cite{lee2013music,cardle2002music}, which shows
limited capacity, while recent researches~\cite{lee2018listen,tang2018dance,lee2019dancing} leverage the deep learning methods to automatically
learn the mapping between music and dance. Usually, prior methods solve music-to-dance synthesis directly by mapping music to human skeleton
keypoints. On account of the highly redundant and noisy nature of skeleton keypoints, the frame by frame keypoints prediction can hardly capture
the coherent structure of music and dance, resulting in limited synthesis quality.

To address this issue, we propose to fuse human choreography knowledge into the music-to-dance synthesis framework. In choreography knowledge,
dance actions are composed of multiple indivisible units. We defined such indivisible unit as choreographic action unit (CAU). A CAU refers to
a clip of human motions that lasts for several musical beats and acts as an undividable unit in choreography. During the choreography procedure
shown in Figure~\ref{fig:motivation}, human choreographers usually devise dance actions in a hierarchical manner, they often recollect CAUs they
have seen or used before and arrange CAUs according to the rhythm, melody and emotion of the music to formulate a piece of dance.

Pondering over such characteristics of choreography, we propose the ChoreoNet, a music-to-dance framework that imitates the hierarchical
choreographic procedure of human beings. Our ChoreoNet has the following two characteristics:
(1) The ChoreoNet firstly applies a CAU prediction model to predict the CAU sequence from musical features. %
Compared with the dance motion prediction in prior works~(predict tens of thousands of frames per music), %
the prediction of CAU sequence~(predict 40\textasciitilde80 CAUs per music) on the one hand prevents the neural network model from learning trival motion details, %
on the other hand lessens the burden of predicting overlong sequences. (2) The ChoreoNet leverages a spatial-temporal inpainting model to convert the CAU
sequence into continuous dance motions. %
Since that there always exists motion gaps between adjacent CAU pairs in the CAU sequence, %
the spatial-temporal inpainting model generates natural transitions between these CAU pairs. %
The overall implementation of ChoreoNet is two-stage, %
during each stage, the corresponding model is trained separately. %

To evaluate the effectiveness of the ChoreoNet framework, we construct a music-to-dance dataset with expert choreographic annotations.
Specifically, we collect 164 CAUs from 4 different dance types~(Cha Cha, Waltz, Rumba and Tango) and record 3D motions of each CAU. %
Then, we collect 62 dance music pieces and invite professional choreographers to annotate the CAU sequence of each music. %
Totally, we have 94 minutes of music with such annotations. Then we perform quantitative and qualitative experiments on our dataset.
Experiments show that compared to baseline methods, our method can generate structured dances with a long duration that match better with
the input music and more natural transitions between adjacent dance motions. Specifically, compared to baseline methods, our framework
generates CAU sequences with higher BLEU score~\cite{DBLP:conf/acl/PapineniRWZ02}. Our motion generation
model also generates a motion transition closer to groundtruth. The dance animation generated by our framework also scores higher in the user study.

To summarize, our contributions are as two-fold: 

\begin{itemize}
  \item We propose to formalize the human choreography knowledge by defining CAU and introduce
  it into music-to-dance synthesis.
  \item We propose a two-stage framework ChoreoNet to implement the music-CAU-skeleton mapping.
  Experiments demonstrate the effectiveness of our method.
\end{itemize}

\section{Related Work}

The previous works related to our ChoreoNet are described in two aspects: music to dance synthesis and human motion generation. %

\subsection{Music to Dance Synthesis}

Several researches have focused on music to dance synthesis. %
Early works usually treat this problem as a mechanical template matching problem~\cite{cardle2002music,shiratori2006dancing,lee2013music}. %
Cardle~\textit{et al.}~\cite{cardle2002music} modified dance motion according to musical features, %
while Shiratori~\textit{et al.}~\cite{shiratori2006dancing} and Lee~\textit{et al.}~\cite{lee2013music} manually defined muscial features and generate dance motions according to musical similarity. %
These template matching methods have limited capacity on generating natural and creative dance motions. %
Later, researchers start to address the dance-to-motions synthesis problem with deep learning techniques~\cite{lee2018listen,tang2018dance,Crnkovic-FriisC16,yalta2019weakly,yaofollow,zhuang2020music2dance,ren2019music}. %
Crnkovic-Friis~\textit{et al.}~\cite{Crnkovic-FriisC16} firstly employ the deep learning methods to generate dance motions, %
they devise a Chor-RNN framework to predict dance motion from raw motion capture data. %
Then, Tang~\textit{et al.}~\cite{tang2018dance} designed a LSTM-autoencoder to generate 3D dance motion. %
Previous research~\cite{ren2019music} also proposed to improve the naturalness of dance motion through perceptual loss~\cite{johnson2016perceptual}. %
However, the redundant and noisy motion keypoints still limit the quality of synthesized dance. %
Yalta~\textit{et al.}~\cite{yalta2019weakly} has proposed to solve such issue through weakly supervised learning, %
whereas, the lack of human choreography experience still makes the generation quality less appealing. %

\subsection{Human Motion Generation}

Human motion generation aims to generate natural human motion conditioned on existing motion capture data and plays a key role in music-to-dance synthesis.
However, the highly dynamic, non-linear and complex nature of human motion makes this task challenging. Early researchers address this problem with
concatenation-based method~\cite{arikan2003motion}, hidden Markov models~\cite{tanco2000realistic} and random forests~\cite{lehrmann2014efficient}.
Then, with the development of deep learning techniques, researchers applied deep learning techniques~\cite{ghosh2017learning,fragkiadaki2015recurrent,
jain2016structural,martinez2017human,DBLP:conf/bmvc/PavlloGA18,hernandez2019human,aksan2019structured} to solve motion generation problems.
Ghosh \textit{el al.}~\cite{ghosh2017learning} and Fragkiadaki \textit{et al.}~\cite{fragkiadaki2015recurrent} focused on RNN-based models with Euler angle
error terms, utilizing the auto-regressive nature of human motion. Later, Pavllo \textit{et al.}~\cite{DBLP:conf/bmvc/PavlloGA18}
addressed the impact of joint angle representation and showed the advantage of quaternion-based representation over Euler angle. Gui \textit{et al.}~\cite{gui2018adversarial}
combined CNN-based model with adversarial training, and achieved better short-term generation quality over previous works. Ruiz \textit{et al.}~\cite{hernandez2019human}
formulated human motion generation as a spatial-temporal inpainting problem and designed a GAN framework to generate large chunks of missing
data. In this work, we devise a similar spatial-temporal inpainting model to generate natural dance motion transitions between adjacent CAUs.

\section{Problem Formulation}
\label{sec:formulate}

% 这一段主要是提出之前formulation的问题，引出我们的formulation。不要长篇大论的再想intro一样写了

Previous researches~\cite{lee2018listen,tang2018dance,Crnkovic-FriisC16,yalta2019weakly,yaofollow,zhuang2020music2dance,ren2019music} have formulated the music-to-dance synthesis as a music-to-keypoint mapping problem. %
However, directly predicting human skeleton keypoints raises a series of difficulties. %
On the one hand, the human skeleton keypoints are usually noisy and redundant, directly mapping music to these noisy keypoints would cause unstable synthesis results. %
On the other hand, each piece of music is usually accompanied by thousands of motion frames, predicting such an overlong sequence is much too challenging for current sequence model.

To address these issues, in this paper, we propose a two-stage music-to-dance synthesis formulation referring to human choreography knowledge. %
Conventionally, dance motions are composed of multiple indivisible units. %
We define such an indivisible units as choreographic action unit~(CAU), details of CAU will be illustrated in section~\ref{sec:cau}. %
During the human choreography procedure, choreographers seldomly improvise dance actions, instead they create dance motion in a two-stage manner: (1) choreographic action unit (CAU) design and (2) CAU sequence arrangement. %
In the first stage, human choreographers often recollect CAUs they have seen or used before. %
Then in the second stage, choreographers arrange CAUs to better fit the rhythm, melody and emotion of a given piece of music. %

% formulation里是不应该讲抽特征这些的
With the observation of such hierarchical choreography procedure, we propose to formulate the music-to-dance synthesis procedure as a two-stage framework. %
Our framework takes music $\mathbf{X}$ as input to generates human skeleton keypoints sequences $\mathbf{C}, \mathbf{C} \in \mathbb{R}^{N \times P}$, where $N$ is the number of motion frames, and $P$ is the number of keypoints. The two stages are formally illustrated as follows: % 

\textbf{CAU Choreography Stage.}
In this stage, given input music $\mathbf{X}$, we aim to generate the corresponding CAU sequence $\{y_1, \dots, y_n\}$, $y_i \in \mathbf{Y}$, where $\mathbf{Y}$ is the overall CAU set. 

\textbf{Motion Generation Stage.}
Having obtained the CAU sequence $\{y_1, \dots, y_n\}$, in this stage, we aim to generate the keypoints sequence $\mathbf{C}$. In this stage, although we have known the keypoints sequence $\mathbf{C}_i$ of particular CAU $y_i$, the transition between adjacent CAUs is unknown. Generating smooth and natural transition between adjacent CAUs is the main problem in this stage.

\section{Choreographic Action Unit Definition}
\label{sec:cau}

\begin{figure}[!t]
  \centering
  \includegraphics[width=0.8\linewidth]{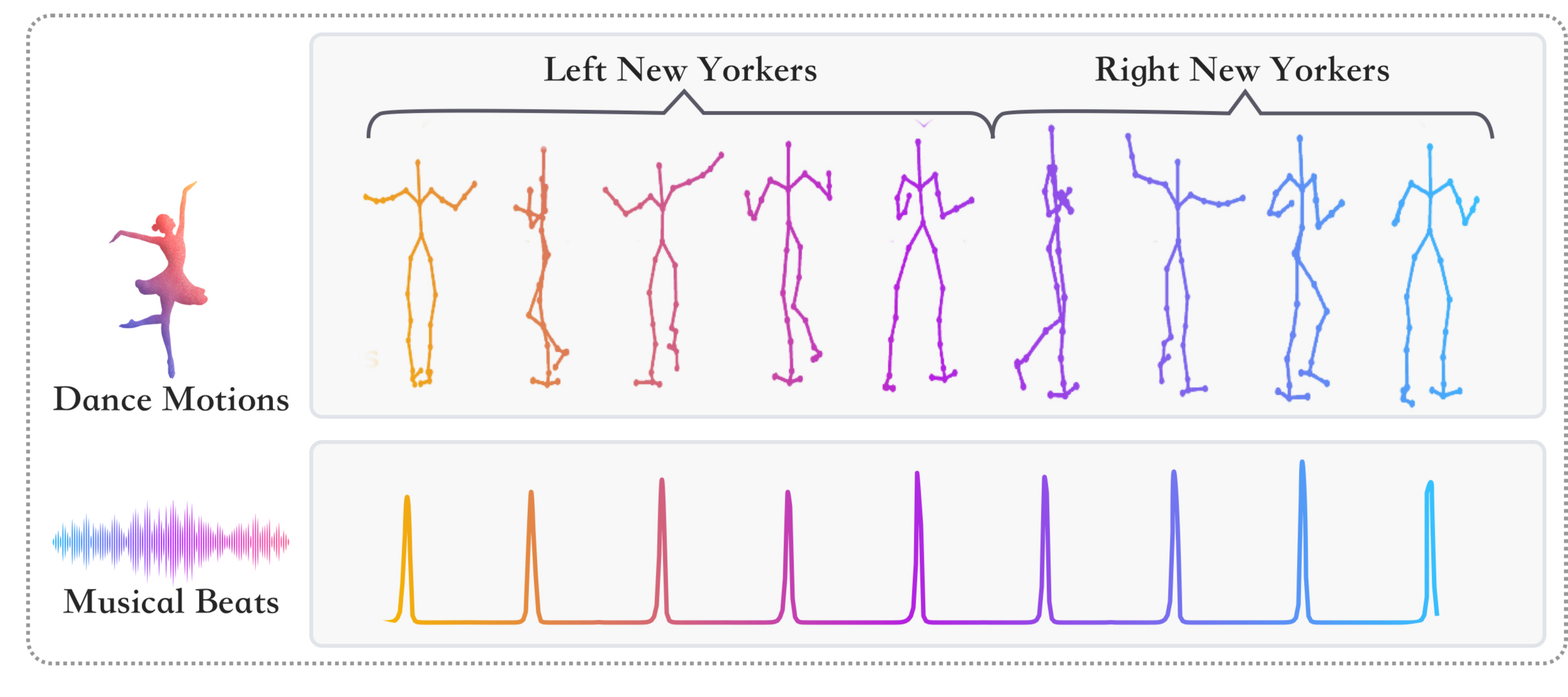}
  \caption{An example of CAUs in an 8 beat motion segment. }
  \label{fig:CAU}
\end{figure}

CAU is short for \textit{choreographic action unit}. One CAU refers to a clip of human motions that lasts for several musical beats and acts as an
undividable unit in choreography. Human choreographers recollect CAUs they have seen or used before and arrange them to formulate new dances.
We ask professional dancers to design CAUs and record their performance of each CAU with motion capture devices. In this work, we denote a CAU
as $y$, $y$ corresponds to $len(y)$ musical beats and $\mathbf{C}^y$ is the corresponding motion capture data performed by professional dancers.

Figure~\ref{fig:CAU} shows the procedure of performing CAUs from a motion segment that lasts for 8 musical beats. At beat 1, the dancer pivots on
the left foot and conducts a quarter turn. At beat 2, the dancer checks forward and spread arms. Afterwards, the dancer moves backward and turn to
the original position at beat 3 and 4. Because the dance motions in the first 4 beats make up a whole action unit and can not be divided into
smaller units in choreography, it is recognized as a CAU named the \textit{Left New Yorkers}. After the \textit{Left New Yorkers}, the dancer
performs the \textit{Right New Yorkers} in the last 4 beats of this motion segment. The dance motions in the last 4 beats are also recognized
as a CAU. So totally we have two CAUs from this motion segment: the \textit{Left New Yorkers} CAU and the \textit{Right New Yorkers} CAU.

\section{Methodology}

\begin{figure*}[!t]
  \centering
  \includegraphics[width=0.6\linewidth]{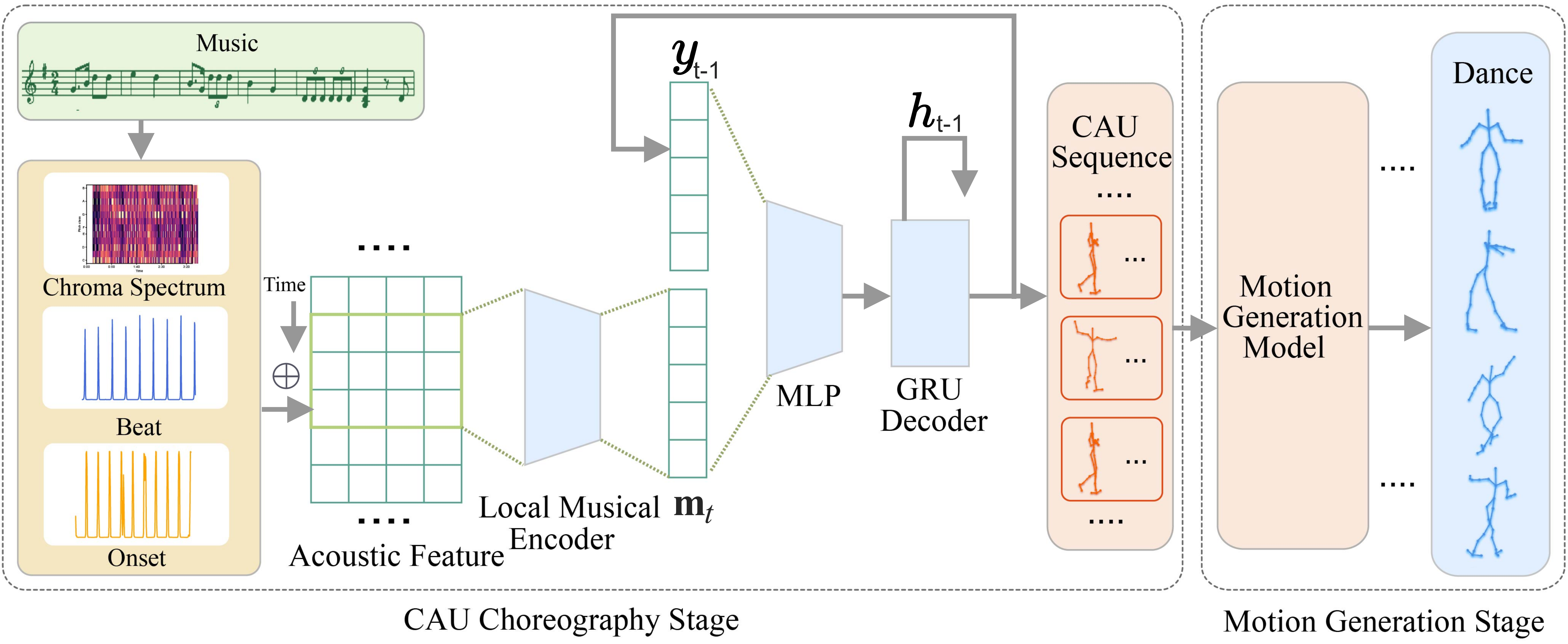}
  \caption{The pipeline of the ChoreoNet. In CAU choreography stage, firstly we extract the deep chroma spectrum, beat and onset of input music. We concatenate the extracted features to formulate acoustic features. Then the local musical encoder computes the encoded musical features $\mathbf{m}_{t}$ in a slide window scheme. The GRU decoder combines $\mathbf{m}_{t}$ and CAU history to predict the next CAU $y_{t}$. In motion generation stage, the motion generation model recovers motion data from CAU sequences. Details of the motion generation model would be illustrated in Figure~\ref{fig:motion generation}. $\oplus$ denotes the concatenation operator. }
  \label{fig:overview}
\end{figure*}

Following the formulation proposed in section~\ref{sec:formulate}, we propose a two-stage framework ChoreoNet, as shown in Figure~\ref{fig:overview}. In the first stage, we generate appropriate choreographic action
unit (CAU) sequences from musical features with an encoder-decoder model. Based on the predicted CAU sequence, in the second stage, we leverage a
spatial-temporal inpainting model to convert CAU sequences to dance motions (human skeleton keypoint coordinates). In the next two subsections,
we will introduce the CAU prediction model and the motion generation model respectively.

\subsection{CAU Prediction Model} % 标题暂时还没写，这段介绍编舞对应的seq2seq

In the first stage of ChoreoNet, we design our CAU prediction model to predict CAU sequence from the input music. As explained above, CAUs should
match with music clips and adjacent CAUs, so our CAU prediction model needs to consider both musical context and CAU context when deciding the next
appropriate CAU. In order to model musical context and CAU context simultaneously, we adopt an encoder-decoder model as the CAU prediction model.
In this model, we first encode musical context with a temporal convolution network, then we feed the encoded musical features to the decoder model.
Our decoder model takes musical features and the CAU history as input and outputs the next CAU. To model the time dependency, we apply gated
recurrent unit (GRU)~\cite{cho2014learning} as our decoder model. The procedure of CAU prediction is shown in Figure~\ref{fig:overview}.
Next, we'll explain each part of the CAU prediction model in detail.

For the encoder of the CAU prediction model, we leverage a temporal CNN to encode music frames. We propose to design the encoder as a local music
encoder (encode a local clip of music) rather than a global encoder (encode the whole input music). This is because the CAU sequence is much sparser
than music frames. One piece of music is usually coupled with only dozens of CAUs but has thousands of music frames. The imbalance between CAUs and
music frames causes difficulty in training a global musicial encoder between the encoder needs much more capacity than the decoder and is hard to
converge. Afterwards, we adopt the following three musical features as the input of the encoder: (1) deep chroma spectrum~\cite{DBLP:conf/ismir/KorzeniowskiW16}
(2) beat~\cite{bock2011enhanced} and (3) onset~\cite{eyben2010universal}. We respectively denote chroma spectrum as $\mathbf{X}^c$, beat as $\mathbf{X}^b$
and onset as $\mathbf{X}^s$. At time $t$, the encoder applies temporal convolution on raw musical features within a fixed-length time window $[t - w, t+ w]$:
\begin{equation}
  \mathbf{m}_t = encode(\mathbf{X}^c_{[t-w, t+w]}, \mathbf{X}^b_{[t-w, t+w]}, \mathbf{X}^s_{[t-w, t+w]}),
\end{equation}
where $\mathbf{m}_t$ represents the encoded musical feature at time $t$.

Having obtained the encoded musical feature $\mathbf{m}_t$, we now model the CAU generation process of the decoder as a probability distribution
conditioned on musical context and CAU context. Given the CAU history $y_1, \dots, y_{t - 1}$ and the encoded musical feature $\mathbf{m}_t$, the
distribution of next CAU $y_t$ at time $t$ is described as:
\begin{equation}
  p(y_t) = p(y_t | {y_1, \dots, y_{t - 1}}, \mathbf{m}_t).
\end{equation}
We model the conditional distribution with a GRU, as shown in Figure~\ref{fig:overview}. $y_{t - 1}$ and $\mathbf{m}_t$ are fused with an MLP
before feeding to the GRU. $\mathbf{m}_t$ represents the musical context, while $y_{t-1}$ and the GRU's hidden state contains the information
of CAU history. In this way, our model has access to both musical context and CAU context.

Combining the two parts of the CAU prediction model, we adopt a sliding-window scheme to generate a CAU sequence from the input music: (1) raw musical features
within the time window $[t - w, t + w]$ is fed to the encoder and we get the encoded musical feature $\mathbf{m}_t$, (2) last predicted CAU $y_p$
and $\mathbf{m}_t$ are fed to the decoder to generate $\hat{y}_t$, (3) $t$ is updated by $t \leftarrow t + len(\hat{y}_t)$. $t$ is initially set to $0$ and the
whole generation process is repeated until an end annotation is generated by the decoder or we meets the end of the input music. Algorithm~\ref{alg:CAU}
shows the whole procedure of the CAU generation.

\begin{algorithm}
  \caption{CAU Generation}
  \label{alg:CAU}
  \begin{algorithmic}[1]
    \State $t \leftarrow 0$
    \State $\mathbf{Y}_{gen} = []$
    \State $y_{p} = StartOfDance$
    \While{$y_{p} \not= EndOfDance$ \textbf{AND} the music is not ended}
      \State $\mathbf{m}_t = encode(\mathbf{X}^c_{[t - w, t + w]}, \mathbf{X}^b_{[t - w, t + w]}, \mathbf{X}^s_{[t - w, t + w]})$
      \State $\hat{y} = decode(\mathbf{m}_t, y_{p})$
      \State Add $\hat{y}$ to $\mathbf{Y}_{gen}$
      \State $y_{p} \leftarrow \hat{y}$
      \State $t \leftarrow t + len(\hat{y})$
    \EndWhile
    \Return $\mathbf{Y}_{gen}$
  \end{algorithmic}
\end{algorithm}

Having obtained the predicted distribution $p(y_t)$ at each step, we train the CAU prediction model by maximizing the log-likelihood of the expert
annotations in the dataset. Given expert annotation $y_t^p$ at time $t$, we minimize the negative log-likelihood loss:
\begin{equation}
  \mathcal{L}_{nll} = -\sum_{t = 1}^{t = N} log(p(y_t = y_t^p)).
\end{equation}

\subsection{Motion Generation Model} % 标题暂时还没写，这段介绍动作拼接

\begin{figure}[!t]
  \centering
  \includegraphics[width=0.85\linewidth]{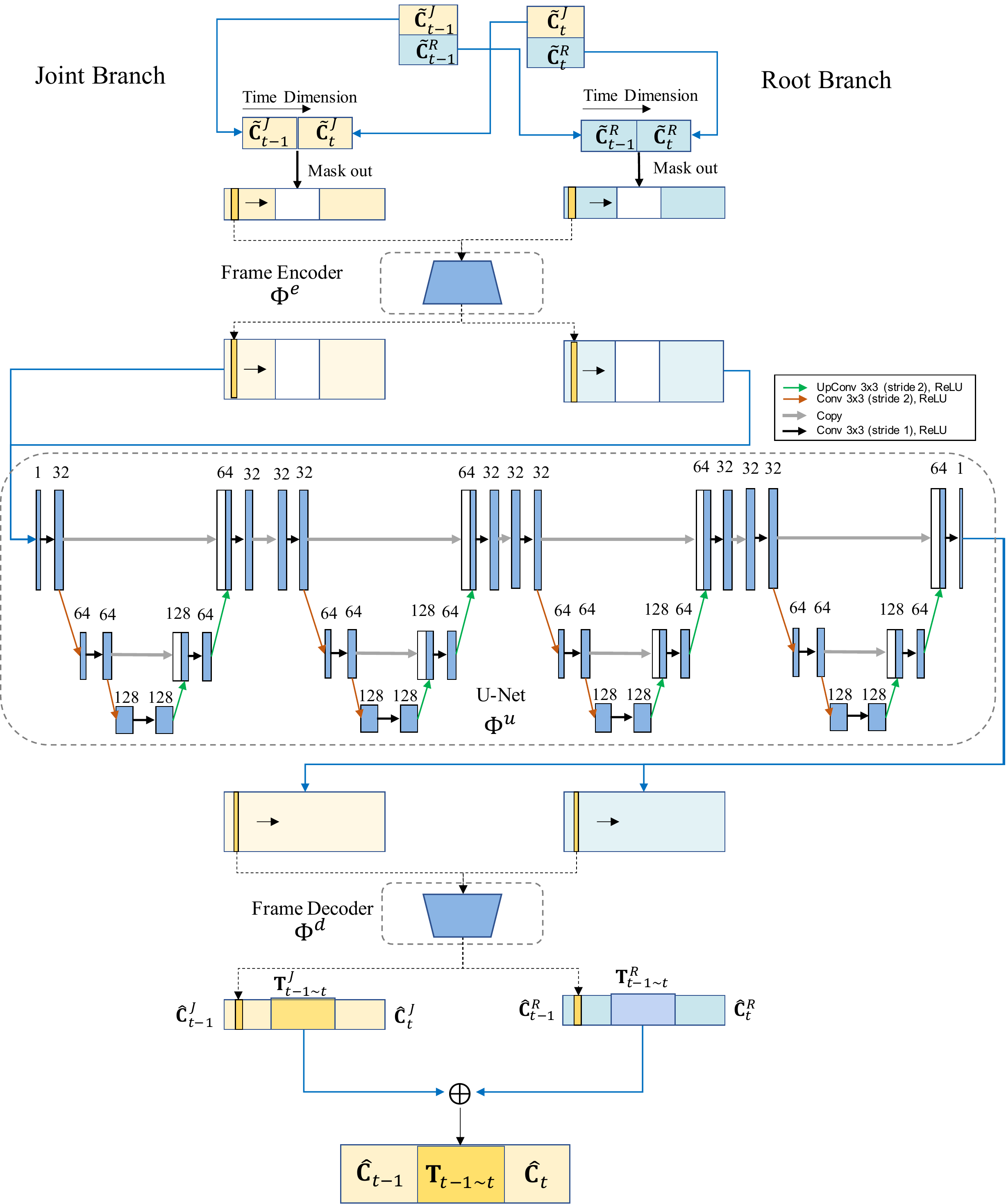}
  \caption{The motion generation model. We generate human joint rotation and root point respectively. The frame encoder firstly projects each frame
  into the latent vector space. Then the U-Net inpaints the transition gap. Finally, the frame decoder projects each frame back into the original
  parameter space. Note that the models of the two branches do not share parameters. }
  \label{fig:motion generation}
\end{figure}

Having obtained the CAU sequence generated from the input music, we convert the CAU sequence to dance motions. Here, dance motions refer to human
skeleton coordinates that can be directly used to drive humanoid animations. For each CAU $y_i$, we collect a clip of motion capture data
$\mathbf{C}_i = \{C_i^1, \dots, C_i^{N_i}\}$ performed by professional dancers, $\mathbf{C}_i$ consists of $N_i$ frames and each pose frame is
denoted as $C_i^j$. Each frame $C_i^j$ is made up of two parts: (1) the rotation of each joint related to its parent joint, (2) the translation and
rotation of the root point (the hip) related to the world coordinate. We adopt quaternion-based joint rotation representation according to Pavllo
\textit{et al.}~\cite{DBLP:conf/bmvc/PavlloGA18}. Thus $C_i^j$ consists of $P = P_r + P_j$ parameters, where $P_r$ represents the number of root
point parameters while $P_j$ represents the number of joint rotation parameters. The motion generation problem is formulated as follows: given
$\{\mathbf{C}_1, \dots, \mathbf{C}_n\}, \mathbf{C}_i \in \mathbb{R}^{N_i \times P}$, we aim to generate $\mathbf{C} \in \mathbb{R}^{N \times P}, N = \sum_{i=1}^n N_i$.

For each clip of motion capture data, a large part of $\mathbf{C}_i$ can be directly used to drive humanoid animation. However,
the motion transition between adjacent CAUs needs to be natural and smooth. To address this issue, we devise a motion generation model to inpaint
dance motions. Before feeding motion capture data to our model, we preprocess $\mathbf{C}_i$ in two steps:
(1) align the root points of adjacent CAUs to avoid sudden change of root position and root rotation, (2) align kinematic beats
(the beats of dance motions) with musical beats. After these steps, $\{\mathbf{C}_1, \dots, \mathbf{C}_n\}$ is converted to
$\widetilde{\mathbf{C}} = \{\widetilde{\mathbf{C}}_1, \dots, \widetilde{\mathbf{C}}_n\}$. Afterwards, we propose to inpaint $\widetilde{\mathbf{C}}$
to generate natural and smooth motion transition between adjacent CAUs. One simple solution could be linear blending.
However, linear blending generates unatural transition when there's huge gap between adjacent CAUs. The reason behind is that
human motion is hightly complicated and non-linear in nature. There exists sudden acceleration, deceleration and body turns in dance motions. To
address this issue, we design a spatial-temporal inpainting model shown in Figure~\ref{fig:motion generation} inspired by Ruiz \textit{et al.}~\cite{hernandez2019human}.

The general idea of spatial-temporal inpainting is similar to image inpainting. Given two clips of motion capture data
$\widetilde{\mathbf{C}}_{t -1}, \widetilde{\mathbf{C}}_t$, the model generates motion transition and its contexts: $\{\hat{\mathbf{C}}_{t - 1}, \mathbf{T}_{t - 1 \thicksim t}, \hat{\mathbf{C}}_t\}$.
$\mathbf{T}_{t - 1 \thicksim t}$ is the generated motion transition, while $\hat{\mathbf{C}}_{t - 1}$ and $\hat{\mathbf{C}}_t$ are predicted contexts.
Usually, $\hat{\mathbf{C}}_{t - 1}$ and $\hat{\mathbf{C}}_t$ are a slightly modified version of $\widetilde{\mathbf{C}}_{t -1}$ and $\widetilde{\mathbf{C}}_t$.
After generating transition for each pair of adjacent CAUs, we get $\mathbf{C}$ by concatenating $\hat{\mathbf{C}}_i$ and $\mathbf{R}_i$.

Our spatial-temporal inpainting model is made up of two sub-models: (1) human joint rotation inpainting model and (2) root point inpainting model. The
two sub-models share the same structure but with different parameters. The workflow of our model is shown in Figure~\ref{fig:motion generation}. The human joint rotation inpainting
model takes only the joint rotation parameters as input, denoted as $\widetilde{\mathbf{C}}_{t - 1}^J$ and $\widetilde{\mathbf{C}}_{t}^J$ in
Figure~\ref{fig:motion generation}. We first concatenate these two matrices and mask out the transition window between the two motion segments
by setting parameters within this window to zeros. Then we projects each frame of the motion segment to a latent vector space $\mathbb{R}^{M}$ through
a frame encoder $\Phi^e$. After that, a 2D U-Net $\Phi^u$ is applied to inpaint the masked  matrix. At last, a frame decoder $\Phi^d$ projects
each frame of the inpainted matrix from $\mathbb{R}^{M}$ to $\mathbb{R}^{J}$, where $J$ is the number of joint rotation parameters. The output of
$\Phi^d$ consists of $\hat{\mathbf{C}}_{t-1}^J$, $\mathbf{T}_{t - 1 \thicksim t}^J$ and $\hat{\mathbf{C}}_t^J$. Similarly, the root point inpainting model takes
the rotation and the velocity of the root point as input, denoted as $\widetilde{\mathbf{C}}_{t - 1}^R$ and $\widetilde{\mathbf{C}}_{t}^R$. The
workflow of the root point inpainting model is the same as that of the joint rotation inpainting model. %
$\hat{\mathbf{C}}_{t-1}$, $\mathbf{T}_{t - 1 \thicksim t}$ and $\hat{\mathbf{C}}_t$ are the concatenation of the outputs from the human joint inpainting model and root inpainting model.

Note that there are two characteristics in the design of our motion generation model. First, we adopt two sub-models for human joint rotation inpainting and
root point inpainting respectively rather than a single model. This is because joint rotation and root velocity are of two different parameter
spaces, it's hard for one inpainting model to operate on two parameter spaces. Second, we project each frame of motion to a larger space of dimension $M$
before feeding the frame into the 2D-UNet. 
That's because 2D-UNet would conduct 2D convolution on the input matrix, %
while the motion frame matrix $\widetilde{\mathbf{C}}_i$, different from the traditional image matrix, is continuous on the time dimension but not on the motion parameter dimension. % 
Thus we cannot directly perform 2D convolution on the frame matrix. To address this issue, we leverages a frame encoder $\Phi^e$ to projects each frame to a motion embedding space, so as to guarantee the continuation property in the embedding space. %
Extensive experiments verify the effectiveness of the two designs.

During training, we adopt the motion capture data as groundtruth. Each time, we randomly clip two consecutive clips of motion capture data denoted as
$\{\mathbf{C}_1, \mathbf{C}_2\}, C_i \in \mathbb{R}^{N_i \times P}$. Then we feed $\{\mathbf{C}_1, \mathbf{C}_2\}$ to the motion
generation model, the human joint rotation inpainting model outputs $\hat{\mathbf{C}}^J = concat(\hat{\mathbf{C}}_1^J, \mathbf{T}_{1 \thicksim 2}^J, \hat{\mathbf{C}}_2^J) \in \mathbb{R}^{(N_1+N_2) \times P_j}$,
the root point inpainting model outputs $\hat{\mathbf{C}}^R = concat(\hat{\mathbf{C}}_1^R, \mathbf{T}_{1 \thicksim 2}^R, \hat{\mathbf{C}}_2^R) \in \mathbb{R}^{(N_1+N_2) \times P_r}$.
Afterwards, we minimize the distance between the outputs and the groundtruth $\mathbf{C}_{gnd} = concat(\mathbf{C}_1, \mathbf{C}_2) \in \mathbb{R}^{(N_1+N_2) \times P}$.
The two sub-models define two different distance function (i.e loss function) respectively: (1) joint rotation loss $\mathcal{L}_{joint}$,
(2) root point loss $\mathcal{L}_{root}$. For joint rotation loss, we adopt geodesic loss proposed by Gui \textit{et al.}~\cite{gui2018adversarial}.
Given two rotation matrices $\mathbf{R}$ and $\hat{\mathbf{R}}$, the geodesic distance is defined using the logarithm map in SO(3):
\begin{equation}
  d_{geo} (\mathbf{R}, \hat{\mathbf{R}}) = ||log(\mathbf{R}\hat{\mathbf{R}}^T)||_2.
\end{equation}
Summing up geodesic distances between the inpainted frames and the groundtruth frames, we obtain joint rotation loss:
\begin{equation}
  \mathcal{L}_{joint} = \sum_{k=1}^{k=N_1+N_2} \sum_{j=1}^{j=J} d_{geo} (\mathbf{R}_k^j, \hat{\mathbf{R}}_k^j),
\end{equation}
where $\mathbf{R}_k^j$ represents the rotation matrix of the j-th joint in the k-th frame of $\mathbf{C}_{gnd}$, while $\hat{\mathbf{R}}_k^j$
represents the rotation matrix of the j-th joint in the k-th frame of $\hat{\mathbf{C}}^J$. For root point loss, we adopt L-1 distance as loss function:
\begin{equation}
  \mathcal{L}_{root} = \sum_{i=1}^{i=N_1+N_2} ||C_i^R - \hat{C}_i^R||,
\end{equation}
where $C_i^R$ is the i-th frame of $\mathbf{C}_{gnd}^R$, while $\hat{C}_i^R$ represents the i-th frame of $\hat{\mathbf{C}}^R$. The two sub-models
are trained separately.

To summarize, we devise a two-stage framework to apply choreography experience into music-dance synthesis. In the first stage, a CAU prediction
model arranges a CAU sequence according to the input music. The CAU prediction model is designed as a CAU prediction model so as to consider both
the musical context and the CAU context when arranging CAUs. In the second stage, we convert the CAU sequence to human motions that can directly
drive humanoid animations. A spatial-temporal inpainting model is used to generate natural and smooth transition between adjacent dance motions.

\section{Experiment}
\label{section:experiment}

In this section, we conduct extensive experiments to demonstrate the effectiveness of our framework. We evaluate our framwork on the CAU annotation
dataset. Our framework has shown better performance compared with baseline methods, both qualitatively and quantitatively. Afterwards, we verify
the effectiveness of our motion generation model with quantitative experiments.

\subsection{Experiment Setup}

\textbf{Dataset.} The expert choreographic annotations we collect come as pairs of music pieces and CAU sequences. Specifically, we collect 62
pieces of dancing music and 164 types of CAUs. For each piece of music, we invite professional choreographers to choreograph for each piece of
music and record their choreography as CAU sequences. We annotate each CAU in the CAU sequece with its start time and end time in the music.
These annotations include four types of dance (Waltz, Tango, Cha Cha, and Rumba) with a total of 94 minutes of music.

The CAU annotations used in this model consist of all the types of CAU we collected from expert choreographers and three other special annotations:
[SOD], [EOD] and [NIL], summing up to a total of 167 annotations. [SOD] and [EOD] represent \textit{start of dance} and \textit{end of dance}
respectively, performing similar functions to those of [SOS] and [EOS] annotations used in NLP problems. [NIL] annotation is introduced to imitate
the decision of start time in choreography. One [NIL] annotation represents waiting for one musical beat. One possible CAU sequence generated by
our model could be: [NIL, NIL, C-1-3, C-18-1]. The first two [NIL] annotations represents that the dance should start after the first two beats,
while [C-1-3] and [C-18-1] are normal CAUs collected from expert choreographers.

Meanwhile, we ask professional dancers to perform dances conssits of the aforementioned CAUs and record their performances through motion capture
devices. Then we crop segments corresponding to each CAU from the motion capture data to form a motion dataset. The FPS (frame per second) of our
motion capture data is 80, we collect a total of 12688 frames of motion capture data. During training, each time, we clip a segment of 192 frames
randomly from the motion capture data and set the motion blending window size to 64.

\textbf{Implementation Details.} We first extract acoustic features from raw input music using \textit{Madmom}~\cite{bock2016madmom} toolkit.
Specifically, we extract deep chroma spectrum~\cite{DBLP:conf/ismir/KorzeniowskiW16} $\mathbf{X}^c$, beat feature~\cite{bock2011enhanced} $\mathbf{X}^b$ and
onset feature~\cite{eyben2010universal} $\mathbf{X}^s$. The frame-per-second (FPS) of $\mathbf{X}^c$ is 10 while the FPS of $\mathbf{X}^b$ and
$\mathbf{X}^s$ is 100. The dimension of each frame of $\mathbf{X}^c$ is 12, while the other two features are both 1D vectors.

Then for the network architecture, the encoder of the CAU prediction model contains 5 1D convolutional layers, and we choose ReLU as activation function.
The sliding window size of this local musical feature encoder is set to 10 seconds. The acoustic features are firstly convolved through the 5
convolutional layers and then fed to an MLP layer to output a 64-dimension local musical feature. The embedding dimension of CAU is set to 128, and
the decoder of the CAU prediction model consists of an MLP layer and a GRU with 64 units.

Before feeding the recorded motion capture segments of the generated CAU sequence to our motion generation model, we align the kinematic beats with
musical beats. A kinematic beat refer to the sudden motion deceleration. We detect the sudden deceleration of human limbs and mark the times of such deceleration as kinematic beats. Afterwards, we align them
with musical beats detected by \textit{Madmom}~\cite{bock2016madmom}.

\begin{figure*}
  \centering
  \includegraphics[width=0.6\linewidth]{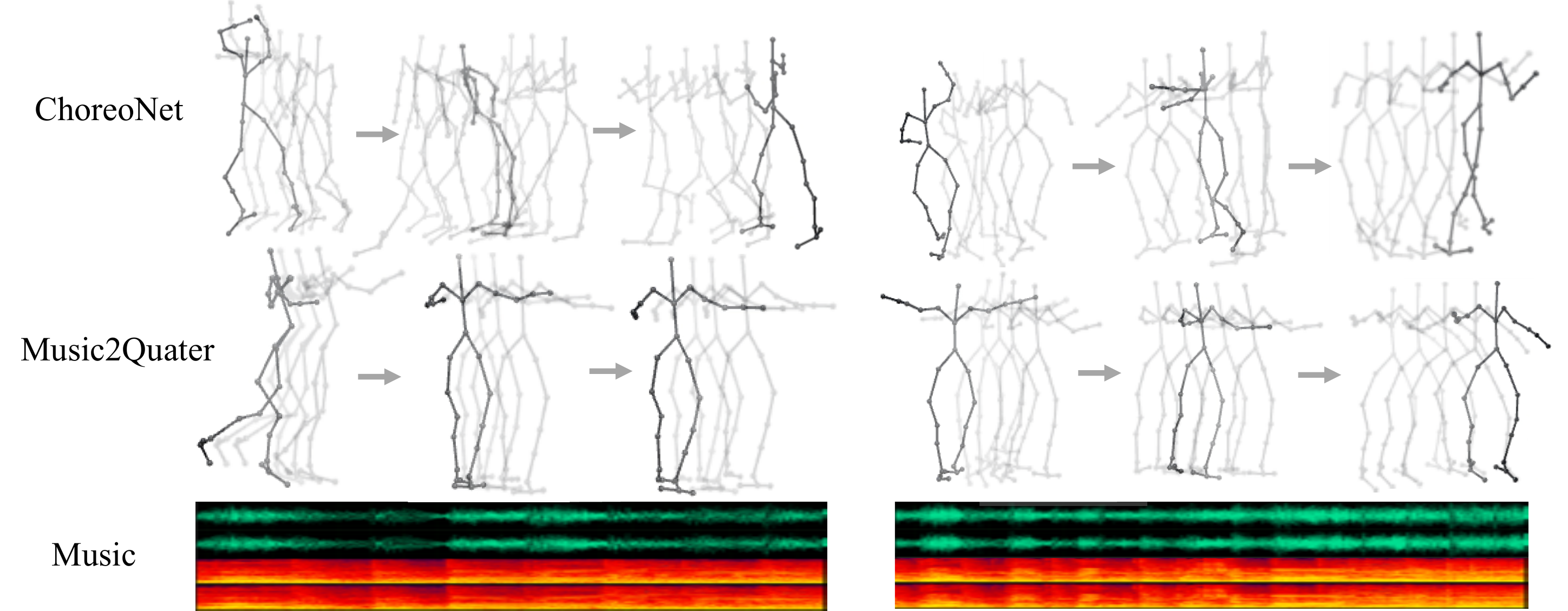}
  \caption{Comparison of Generated Dances. The dances generated by Music2Quater (directly music-to-dance mapping baseline) collapse to a static pose
  after the first few seconds of music, while our ChoreoNet can generate different motions according to the input music. }
\end{figure*}

\begin{table*}
\caption{Comparison of Different Methods}
  \begin{tabular}{ccccc}
    \toprule
    Methods & BLEU-4 & Music Matchness & Motion Naturalness & Motion Multimodality\\
    \midrule
    Music2Quater & - & 2.47 & 2.31 & 2.09 \\
    One-to-one CAU Prediction & 0.0823 & - & - & - \\
    ChoreoNet Non-recurrent & 0.539 & 3.78 & \textbf{4.05} & 3.80 \\
    ChoreoNet & \textbf{0.704} & \textbf{3.88} & 3.94 & \textbf{3.82} \\
    \bottomrule
  \end{tabular}
  \label{tab:ChoreoNet Performance}
\end{table*}

The frame encoder $\Phi^e$ and the frame decoder $\Phi^d$ of the motion generation model are both 1D convolutional networks with 6 convolutional
layers and we adopt ReLU as their activation function. The dimension of motion embedding space is set to 84. The 2D U-Net $\Phi^u$ consists of
4 blocks, each block is made up of 4 downsample layers and 4 upsample layers. The number of hidden states of each block is set to 32. In practice,
we mask out the middle part of a clip of motion capture data and then feed the masked clip to the motion generation model to reconstruct the original
motion. The size of $\mathbf{G}_i$ are set to 64 frames, i.e. the model takes 192 frames of motion data and outputs 192 frames of inapinted motion
data.

We adopt RMSprop algorithm~\cite{graves2013generating} to train our CAU prediction model for 1000 epochs with an initial learning rate at $10^{-3}$.
We apply \textit{ReduceOnPlateu} learning rate decay strategy provided by PyTorch~\cite{DBLP:conf/nips/PaszkeGMLBCKLGA19} with the patience set to
8 and the decay factor set to 0.9. For the motion generation model, we adopt Adam algorithm~\cite{DBLP:journals/corr/KingmaB14} as optimizer and
train the model for 400 epochs with a mini-batch size of 48 samples and a initial learning rate at $10^{-3}$.  \textit{ReduceOnPlateu} strategy is
also applied, with the patience set to 5 and the decay factor set to 0.7.

\subsection{Metrics}

To evaluate the effectiveness of our framework, we adopt different metrics for the CAU prediction model and the motion generation model respectively.
For the CAU prediction model, BLEU score~\cite{DBLP:conf/acl/PapineniRWZ02} is used for quantitative comparison. BLEU score is originally designed to
evaluate the quality of machine-translated text. In our experiment, we adopt it to evaluate the quality of the generated CAU sequences. For the motion
generation model, we evaluate the geodesic distance and FID~\cite{DBLP:conf/nips/HeuselRUNH17}. For geodesic distance, we evaluate the human joint
rotation distance between the generated transition and the groundtruth. For FID, we evaluate the distance between the feature of the generated human
motions and the feature of the real human motions. We train a motion auto-encoder on our motion capture dataset as the feature extractor for our
FID evaluation.

We also conduct a user study to evaluate the quality of the dances generated by out framework. Participants are required to rate the following
factors from 0 to 5: (1) the matchness between dance and music, (2) the naturalness of the dance motions, (3) the multimodality of the dances.

% qualitative evaluations
% (放一些连续帧的图，以及user study)

\subsection{Comparison with Baselines}
% comparison with baselines

In this section, we compare the following baseline methods: (1) Music2Quater. This method directly maps musical features
to human skeleton keypoints. It's similar to the LSTM-autoencoder proposed by Tang \textit{et al.}~\cite{tang2018dance}, but replaces LSTM with GRU and
adopts quaternion as motion representation. (2) One-to-one CAU predition. This method takes musical features as input and outputs CAU sequences without
refering to the previous CAU context. It's implemented by replacing the decoder of our CAU prediction model with a simple MLP layer. (3) ChoreoNet 
non-recurrent. This method replaces the decoder of our CAU prediction model with a CNN and takes the last predicted CAU as input. Because the
new decoder does not have hidden state, it's only capable to capture the local CAU context rather than the whole CAU context. (4) The proposed
ChoreoNet framework.

% quantitative evaluations
% 这里不要写拼接的实验了，拼接的在下一章写，只是写整体的结果
% 可以参考https://arxiv.org/pdf/2002.03761.pdf
% 也可以放机翻的指标(因为music2points可能没法参考机翻指标，表格列出来，那里写一个-就行)

\textbf{Quantitative Evaluations.} We perform a quantitative evaluation of the quality of the CAU sequences generated from the input music. Specifically,
we select 5 pieces of dancing music that are not in our training dataset. Afterwards, we evaluate the BLEU scores of CAU sequences generated from
these music. The BLEU score measures the similarity between the generated CAU sequences and the CAU sequences we collected from professional
choreographers. Table~\ref{tab:ChoreoNet Performance} shows the evaluation results. The BLEU score of Music2Quater is not given because it does
not produce CAU sequences in the dance motion synthesis process. Overall, we can see the CAU context is crucial to CAU prediction. The one-to-one
CAU prediction method ignores CAU context produces poor results. Replacing the GRU decoder with a CNN would also cause performance degradation.
Our framework outperforms the ChoreoNet non-recurrent baseline by 0.165 with $0.704$ BLEU score on our CAU annotation dataset.

\textbf{Qualitative Evaluations.} We compare the quality of dances synthesized by different methods. Dances generated by Music2Quater tend to collapse
to a static pose after first few seconds. The noisy and highly redundant nature causes difficulty for the model to directly map musical feature
to human skeleton keypoints. For ChoreoNet w/o CAU context, the generated dances are made up of some random actions, while ChoreoNet with local CAU
context is able to generate valid dances but tend to produce some repeated actions. This proves that CAU context is crucial to our framework. To
further investigate the quality of synthesized dances, we also conduct a user study. 17 participants are asked to rate 'matchness with music', 'motion
naturalness' and 'motion multimodality' of each synthesized dances. The results of the user study is shown in Table~\ref{tab:ChoreoNet Performance}.
From the results, we can see that compared to the Music2Quater baseline, our ChoreoNet framework scores 1.41 points higher in \textit{Music Matchness},
1.63 points higher in \textit{Motion Naturalness} and 1.73 points higher in \textit{Motion Multimodality}. The results confirm our observation
that the ChoreoNet framework generates dances of higher quality than directly music-to-skeleton mapping methods.

\subsection{Analysis on Motion Generation Model}
% analysis on choreonet
% 这里放拼接模型的分析
% 对各个参数的分析(窗口大小之类的)
% ablation study
% 全部是定量实验

In order to evaluate the performance of our spatial-temporal inpainting model, in this section, we study it from three aspects. First, we compare it
with linear blending baseline. Next, we conduct ablation study to verify the effectiveness of the frame encoder/decoder and the two sub-model design.
Finally, we adjust the blending window size to analyze the impact of different blending window sizes.

\textbf{Comparison with Baseline.} To verify the effectiveness of our spatial-temporal inpainting model, we compare it with the linear blending
baseline. This is a solution used by previous researchers to concatenate two adjacent segments of motion data. Specifically, we conduct linear
blending on human joint rotation within the blending window. The results are shown in Table~\ref{tab:lb}. Our method achieves produce motion
transition with much smaller geodesic distances than the baseline.

\textbf{Ablation Study.} To verify the effectiveness of our spatial-temporal inpainting model, we compare it with the following methods. (1) Ours
w/o sub-model. This method ablates the two sub-model design, there is only one model to inpaint both human joint rotation and the root point.
(2) Ours w/o frame encoder. This method ablates the frame encoder and decoder, all the motion inpainting is conducted in the original motion
parameter space. We calculate the geodesic distance per frame between the generated motion transition and the groundtruth to evaluate the
effectiveness of these methods.

\begin{table}
  \caption{Comparison with Linear Blending. We evaluate the geodesic distances and FID of the linear blending baseline and our spatial-temporal
  inpainting method with different blending window sizes. }
  \label{tab:lb}
  \begin{tabular}{ccc}
    \toprule
    Methods & Geodesic Distance & FID \\
    \midrule
    Linear Blending (16 window) & $1.72 \times 10^{-2}$ & $85.1$ \\
    Ours (16 window) & $1.24 \times 10^{-3}$ & $76.9$ \\
    Linear Blending (32 window) & $1.80 \times 10^{-2}$ & $75.8$ \\
    Ours (32 window) & $1.31 \times 10 {-3}$ & $67.8$ \\
    Linear Blending (64 window) & $1.95 \times 10^{-2}$ & $69.7$ \\
    Ours (64 window) & $\mathbf{1.33 \times 10^{-3}}$ & $\mathbf{63.9}$ \\
    Linear Blending (128 window) & $2.08 \times 10^{-2}$ & $105.4$ \\
    Ours (128 window) & $1.97 \times 10^{-3}$ & $78.4$ \\
    \bottomrule
  \end{tabular}
\end{table}

\begin{table}
\caption{Comparison with Ours w/o frame encoder and Ours w/o sub-model. The blending window size is set to 64.}
  \label{tab:ablation}
  \begin{tabular}{cc}
    \toprule
    Methods & Geodesic Distance \\
    \midrule
    Ours w/o frame encoder & $2.02 \times 10^{-3}$ \\
    Ours w/o sub-model & $5.14 \times 10^{-3}$ \\
    Ours & $\mathbf{1.33 \times 10^{-3}}$ \\
    \bottomrule
  \end{tabular}
\end{table}

Table~\ref{tab:ablation} shows the test geodesic distances of all the methods. The results show that the designs of frame encoder/decoder and sub-models
increase the performance by a large margin.

\textbf{Motion Blending Window Size Analysis.} The size of motion blending window is an important hyper-parameter. Small blending window would cause
quick and unatural motion transition while big blending window causes difficulty to our model as it needs more capacity to inpaint a larger segment
of missing motion transition. The same holds true for the linear blending baseline method. As it has no other parameters, both big or small blending window would cause undesirable results.
To evaluate the quality of generated motion transition, we adopt FID~\cite{DBLP:conf/nips/HeuselRUNH17}
to measure the distance between the generated motion transition and the real motion capture data. As there exists no standard feature extractor for
motion data, we train a motion auto-encoder on our motion capture dataset as motion feature extractor.

From the results shown in Table~\ref{tab:lb}, we observe that the minimum FID of our model occurs at blending window size of 64 frames. Both small
transition window size and large transition window size increase the FID. The results confirm the observation that proper transition window size
leads to better transition quality.

\section{Conclusion}

In this paper, we formulate music-to-dance synthesis as a two-stage procedure to introduce human choreography experience. Firstly, we define choreographic
action unit (CAU) and build a dataset containing 62 pieces of dancing music and 164 types of CAUs. Each piece of music is coupled with human
expert CAU annotations. Based on the dataset, we propose a music-to-dance synthesis framework to implement the two-stage music-to-dance synthesis
procedure. In the first stage, a CAU prediction model is used to generate CAU sequences from musical features. Then we apply a spatial-temporal
inpainting model to generate dance motions from CAU sequeces. We conduct extensive experiments to verify the effectiveness of our framework. The results
show that compared to baseline methods, our CAU prediction model generate CAU sequeces of higher quality and our spatial-temporal inpainting model
produce more natural and smoother motion transition.

Overall, our framework improves the music-to-dance synthesis performance by a large margin. Furtherly, the proposed CAU-based formulation paves a
new way for future research.

\section*{Acknowledgements}
This work is supported by National Key R\&D Program of China (2019AAA0105200), the state key program of the National Natural Science Foundation of China (NSFC)
(No.61831022), and the National Natural, and Science Foundation of China (61521002). We also want to thank Tiangong Institute for Intelligent Computing,
Tsinghua University and Beijing Academy of Artificial Intelligence (BAAI) for their support.

\newpage

\bibliographystyle{ACM-Reference-Format}
\bibliography{base}

\appendix

\end{document}